\documentclass[10pt,twocolumn,letterpaper]{article}
\usepackage[final]{cvpr} 
\def\cvprPaperID{21} %
\def\confName{CVPR}
\def\confYear{2023}

\usepackage{hyperref} 
\usepackage{mathrsfs}
\usepackage{amsfonts}
\usepackage{amsmath}
\usepackage[flushleft]{threeparttable}
\usepackage{tablefootnote}
\usepackage{multirow}
\usepackage{graphicx}

\usepackage{enumitem}
\usepackage{hanging}
\usepackage{color}
\usepackage{bigstrut}
\usepackage{tikz}
\usetikzlibrary{calc,arrows,decorations.markings}
\usepackage{balance}
\usepackage{tabularx}

\newcommand{\object}{o}
\newcommand{\pose}{P}
\newcommand{\volume}{V}
\newcommand{\feature}{\mathbf{f}}
\newcommand{\scene}{S}

\newcommand{\nerfmlp}{\Psi}
\newcommand{\graspmlp}{\Phi}

\newcommand{\camerapose}{P^{cam}}
\newcommand{\ray}{\vec{r}}

\graphicspath{{../},{../image}}

\usepackage{booktabs}  %

\graphicspath{{../},{../image}}

\usepackage{booktabs}  %

\usepackage{cuted}
\usepackage{capt-of}

\title{\LARGE \bf One-Shot Neural Fields for 3D Object Understanding}

\author{Author Names Omitted for Anonymous Review. Paper-ID 84}
\author{Valts Blukis$^{1}$, Taeyeop Lee$^{1, 2}$, Jonathan Tremblay$^{1}$, Bowen Wen$^{1}$, In So Kweon$^{2}$, \\
 Kuk-Jin Yoon$^{2}$, Dieter Fox$^{1}$, Stan Birchfield$^{1}$\\
$^{1}$NVIDIA, $^{2}$KAIST \\
}

\begin{document}

\maketitle

\begin{abstract}
We present a unified and compact scene representation for robotics, where each object in the scene is depicted by a latent code capturing geometry and appearance.
This representation can be decoded for various tasks such as novel view rendering, 3D reconstruction ({\em e.g.}, recovering depth, point clouds, or voxel maps), collision checking, and stable grasp prediction.
We build our representation from a single RGB input image at test time by leveraging recent advances 
in Neural Radiance Fields (NeRF) that learn category-level priors on large multiview datasets,
then fine-tune on novel objects from one or few views.
We expand the NeRF model for additional grasp outputs and explore ways to leverage this representation for robotics.
At test-time, we build the representation from a single RGB input image observing the scene from only one viewpoint.
We find that the recovered representation allows rendering from novel views, including of occluded object parts, and also for predicting successful stable grasps.
Grasp poses can be directly decoded from our latent representation with an implicit grasp decoder.
We experimented in both simulation and real world and demonstrated the capability for robust robotic grasping using such compact representation.
Website: \url{https://nerfgrasp.github.io/}.

\end{abstract}

\section{Introduction}

Methods that represent scenes with implicit functions of physical quantities have swept through the computer vision and graphics communities, with Neural Radiance Fields (NeRF)~\cite{mildenhall2020nerf} arguably the most prominent example; 
they are also beginning to %
impact state representation in robotics, and are already used successfully 
for grasping~\cite{kerr2022evonerf, yen2022nerfsupervision,lin2022mira,IchnowskiAvigal2021DexNeRF,dai2022graspnerf},
rendering tactile images~\cite{zhong2022touching},
learning multi-object dynamics for manipulation~\cite{driess2022learning},
robot reinforcement learning~\cite{driessnerf},
camera pose estimation~\cite{yen2020inerf}, 
dynamic control~\cite{li2021visuomotor},
planning~\cite{adamkiewicz2022vision},
scene reconstruction~\cite{imapSucar:etal:ICCV2021}, 
and object reconstruction~\cite{abou2022implicit}.
These methods provide a glimpse into the future of state acquisition for robotics applications, but most come with some caveats:
they can be slow to train, or need multiple image acquisitions at test-time;
the final representation is likely to be inflexible (needing retraining if the scene arrangement changes), 
or the scene representation might be hard to reason about spatially, {\em e.g.}, hash tables~\cite{mueller2022instant}. 

\begin{figure}[!t]
    \centering
    \includegraphics[width=0.5\textwidth]{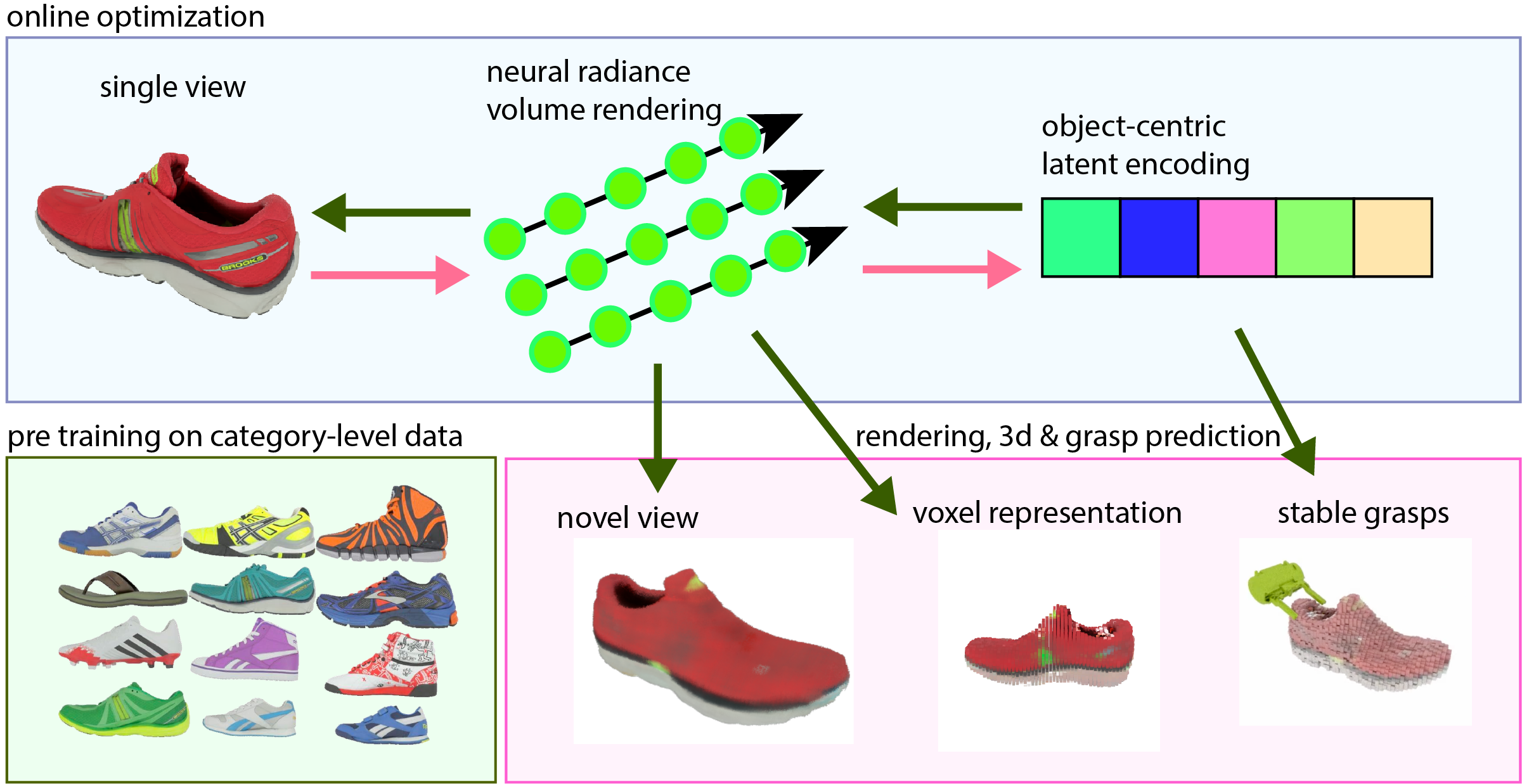}
    \caption{We present a category-level scene representation for robotics applications. 
    Our system is pre-trained on a set of synthetic data (green box). 
    At runtime, we perform an auto-decoding optimization through a neural radiance field to recover a latent code for each object from a single RGB input view (blue box).
    Once the optimization converges to a latent representation, we can render the object in a novel view, generate stable grasps, and reason about the object's 3D shape to generate occupancy maps or query collisions, all from the latent representation.  
    The red arrows represent optimization path to update the latent representation.
    The green arrows represent generation of various outputs from the representation.
    }
    \label{fig:overview_system}
\end{figure}

We introduce a scene representation that expresses each object in the scene with a latent vector that encodes its geometry and appearance, alongside a pose and a bounding volume, see Figure~\ref{fig:overview_system} for an overview.
A neural network 
is used to decode this latent space into two implicit field quantities: a neural radiance field and a neural grasp field.
The radiance field enables synthesizing novel views of the scene including of occluded and unobserved parts of the objects, as well as recovering 3D reconstructions with shape completion.
The grasp field is used for proposing stable and collision-free grasps for each object.
One of our main insights is that a latent representation of an object obtained via inverse volumetric rendering of a pre-trained NeRF, is also useful for predicting other quantities relevant to robotics.
In this paper, we show that from a single partial-view RGB image observed on a real robot at test-time, we can recover a representation of previously unseen objects that enables collision-free stable grasp generation, 3D reconstruction, and generating novel views.
The decoder is pre-trained for image reconstruction and grasp prediction on a category-level multiview dataset of similar virtual object models.
This pre-training design is related to NeRF methods for few-view novel image synthesis~\cite{yu2021pixelnerf, mueller2022autorf}, and also to grasp learning methods~\cite{sundermeyer2021contact}, which we expand to train a continuous neural field.
One of our design goals is simplification of robotics systems around a unified representation, therefore we focus on a single representation useful for a variety of tasks while remaining compact and fast to optimize, rather than excelling at any particular task.

We also propose an original dataset of synthetic shoe objects, high-quality multi-view renders for NeRF training, and grasp annotations. 
We evaluate our method quantitatively on both grasp quality and reconstruction quality, and we compare it against strong baselines. 
Finally, we present qualitative results on a real robot, showing that our system can be used to grasp physical objects
when integrated with modern computer vision systems, namely CenterPose~\cite{lin2022icracenterpose} for object pose estimation and ViLD for open-vocabulary segmentation~\cite{gu2022vild}.
Even though our system was pre-trained on a dataset of synthetic objects, we find that it can be deployed on real robots without sim2real training or accounting for the domain gap.

\section{Related Work}

Robotics perception systems are grounded in 3D understanding.
In recent years 
we have seen state acquisition systems for robotics expand from  
RGB-only pose estimation methods for known objects~\cite{tremblay2018deep,labbe2022megapose} 
to point-clouds used for grasping unknown objects~\cite{mousavian2019graspnet},
to tactile images~\cite{zhong2022touching}, 
to unknown object pose estimation~\cite{lin2022icracenterpose,labbe2022megapose,lee2021category,wen2021bundletrack}.
These perception systems use various representations such as poses, volumes, depth images, point clouds, 
{\em etc}.
These methods
have different trade-offs in their representation of the world state. 
For example, 
a pose does not depict the size of an object, while a depth map does not 
capture any information in occluded regions. 
These trade-offs limit the ability to combine multiple advanced 
applications 
in robotics within a single system, 
such as grasping unknown objects/categories, language grounding, scene editing, 
planning, robot motion generation, {\em etc}. 

Recently neural radiance fields (NeRFs) \cite{mildenhall2020nerf}
have emerged as a popular scene representation in computer vision and graphics. 
NeRF is a function parameterized by a neural network that maps a coordinate and direction vector in Euclidean space to two values: color and density. 
A NeRF can be queried on coordinates along rays to render an image of the scene from any viewpoint. 
These approaches are promising due to their ability to represent the geometry and appearance of a scene using a fixed parameter budget, 
although training can be time consuming and data hungry.
Recent works have explored different approaches to accelerate NeRF training:
instant-NGP uses a hash grid and CUDA acceleration~\cite{mueller2022instant},
while Plenoxels uses a voxel based method to accelerate both training and rendering~\cite{yu_and_fridovichkeil2021plenoxels}.

PixelNeRF is one of the first methods to address few-shot novel view synthesis. %
It trains a large category-level model on multiple scenes~\cite{yu2021pixelnerf}. 
At test-time the method consumes a small number of images and camera poses and learns a grid representation 
that can be used to render the scene from new perspectives. 
EG3D~\cite{Chan2022eg3d} builds on this to use generative-adversarial training, and represents objects with latent codes that can be fitted from a single image by pivotal tuning inversion.
3D representations from few images have also been explored for 
producing textured meshes of zebras and birds~\cite{li2020eccv:sssv3d}.  

We took inspiration from object focused NeRF methods, such as CodeNeRF~\cite{jang2021codenerf}, and AutoRF~\cite{mueller2022autorf}.
Both methods represent a radiance field with multiple per-object latent codes.
CodeNeRF requires tens of images at test-time, while AutoRF uses pre-trained models and can function from fewer test-time images.
Like AutoRF, our proposed method also represents a complex scene by assigning each object a latent code.
Whereas we find these codes through optimization, AutoRF focuses on an image encoder approach.
Other impressive methods have explored editing of object-centric NeRFs, such as \cite{yang2021objectnerf}.

\begin{figure*}[h]
    \centering
    {\includegraphics[width=\textwidth,clip=true,trim=0in 0in 0in 0in]{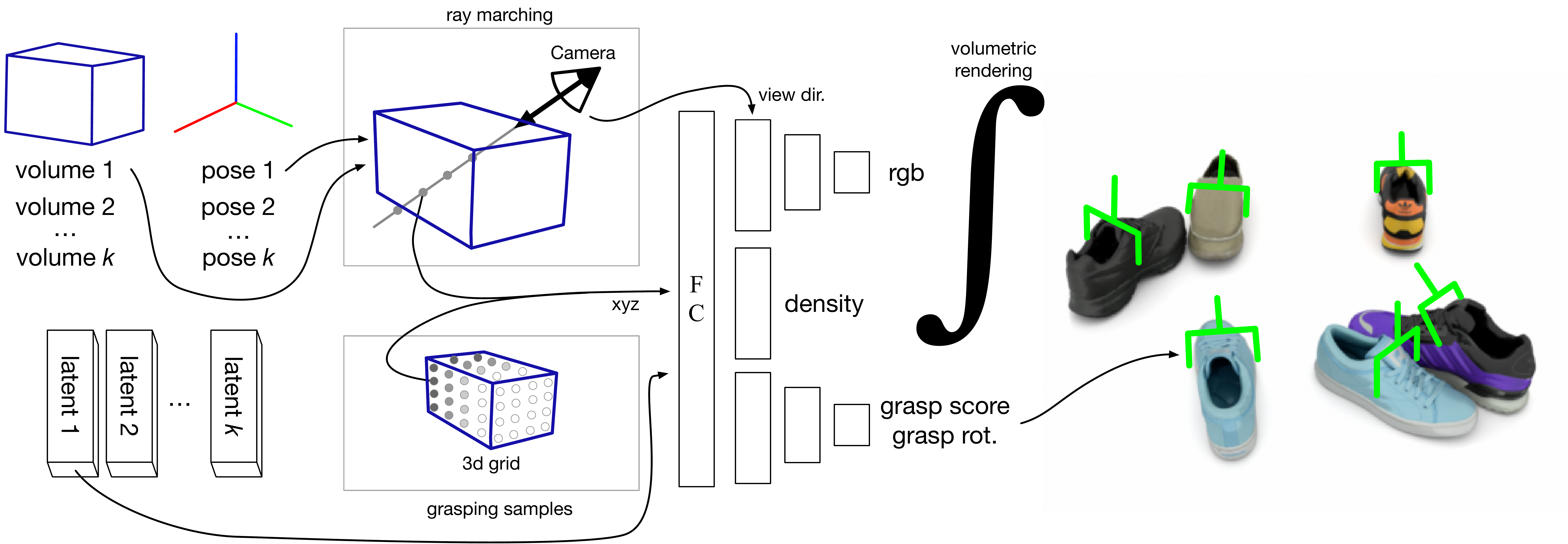}} 
    \vspace{-2em}
    \caption{
    We present a unified representation that can be used to re-render scenes and generate grasping poses.
    Given object poses (top left), volumes (top left), and learned latent codes (bottom left), 
    the system renders the scene in novel configurations and annotates each object with a grasping pose (right image). 
    The rendering ray-marches through a set of 
    volumes (cuboids in which the objects reside). 
    In order to generate grasps (shown in green on the right), 
    we use a grid sampling approach that covers each object volume. 
    Grasp orientation and score are predicted using a decoder which is partially shared with the volumetric rendering process.
    We select the grasp point that yields the highest confidence score.
    FC refers to a fully connected layer. \label{fig:abstract_demo}
    }
    \label{fig:nerf}
\end{figure*}%

NeRF methods have also been applied to robotics, 
in reinforcement learning~\cite{driessnerf}, 
for robot planning ~\cite{pantic2022sampling,adamkiewicz2022vision,turki2022mega,driess2022learning,zhou2023nerf},
and to find depth maps from implicit functions for both thin objects~\cite{yen2022nerfsupervision} and transparent objects~\cite{IchnowskiAvigal2021DexNeRF}.

Robots have also been used by the graphics and computer vision community to generate interesting 
motion problems to solve \cite{noguchi2022watch,li2021visuomotor}.
While impressive, many of these methods suffer from difficulty in generalising to new scenes, 
tedious capturing procedures, long training time, and/or long rendering time. 
We believe that our proposed method alleviates these concerns.
Recently, Evo-NeRF~\cite{kerr2022evonerf} and GraspNeRF~\cite{dai2022graspnerf} presented an exciting new method 
that can quickly edit an already-trained NeRF representation to take into consideration 
objects that have been grasped. 
They also presented a method that maps depth images rendered from NeRF  to produce grasp poses on transparent objects.
In contrast, our method generates grasps directly from object latent codes, avoiding the need for separately trained downstream models, and requiring less input views.

Several methods have been proposed to retrieve grasp poses on objects. 
GraspNet~\cite{mousavian2019graspnet} processes a point cloud using an auto-encoder to generate possible 
object grasps. Jiang {\em et al.}~\cite{jiang2021rss:syn} used point clouds to generate implicit functions from which they optimize grasp points.
Our proposed method for finding grasp poses through a grid search is similar to the concurrent 
work of Shridhar {\em et al.}~\cite{shridhar2022peract}, in which gripper poses are sampled on a voxel grid to find the correct next action.
Category-level neural representations have been explored to allow robots to re-execute demonstrations in different 
configurations~\cite{wen2022you,simeonov2022icra:ndf}.
These works generally aim to develop narrowly focused representations that excel on the particular task. In contrast, we aim to develop a universal representation suitable for a range of tasks.

\section{Model}

In this section we first define our scene representation, then describe our compositional nerf rendering approach, and finally we describe our grasp proposal approach.
Figure~\ref{fig:abstract_demo} shows an overview of information flow in our method.

\subsection{Scene Representation}

\newcommand{\pos}{\vec{p}}
\newcommand{\dir}{\vec{\omega}}
\newcommand{\depth}{d}
\newcommand{\numsteps}{J}
\newcommand{\stepidx}{j}
\newcommand{\numobjs}{M}
\newcommand{\objidx}{m}
\newcommand{\numrays}{N}
\newcommand{\rayidx}{n}

\newcommand{\render}{\textsc{Render}}
\newcommand{\graspinference}{\textsc{PredictGrasps}}

We represent each of the $\numobjs$ objects in the scene $\scene$ with a tuple $\object = (\pose_{\object}, \volume_{\object}, \feature_{\object})$, where $\pose_{\object}$ is a 6D object pose, $\volume_{\object}$ is a 3D object bounding volume, and $\feature_{\object}$ is a learned object latent code.
$\feature_{\object}$ is intended to encode the shape, appearance, and grasp poses of the object.
$\volume_{\object}$ is a simple volume such as a sphere or an oriented box to support fast ray-volume intersection checks.
This representation affords direct control over individual object poses to support visualizing or reasoning about the scene under object rearrangement.
It also fits naturally with the computer vision systems which we leverage for the detection and tracking of objects.

The per-object latent code
can be fed into neural network based decoders for downstream tasks.
In this work we study a neural radiance field decoder~$\nerfmlp$ and a grasp field decoder~$\graspmlp$.
The two decoders share the parameters of their first two layers.
The neural radiance field decoder enables novel view synthesis of rgb or depth images via volumetric rendering, reconstruction with shape completion, and object-point collision checks.
The grasp field decoder enables predicting 6-DoF grasps proposals for robot manipulation.
This representation is motivated by the ability to add additional decoders in the future, for example for object-object collision checking or dynamics modelling~\cite{driess2022learning}.

\subsection{Volumetric Rendering with the Radiance Field Decoder}

\begin{figure}
    \centering
    \scriptsize
    \includegraphics[width=0.5\textwidth]{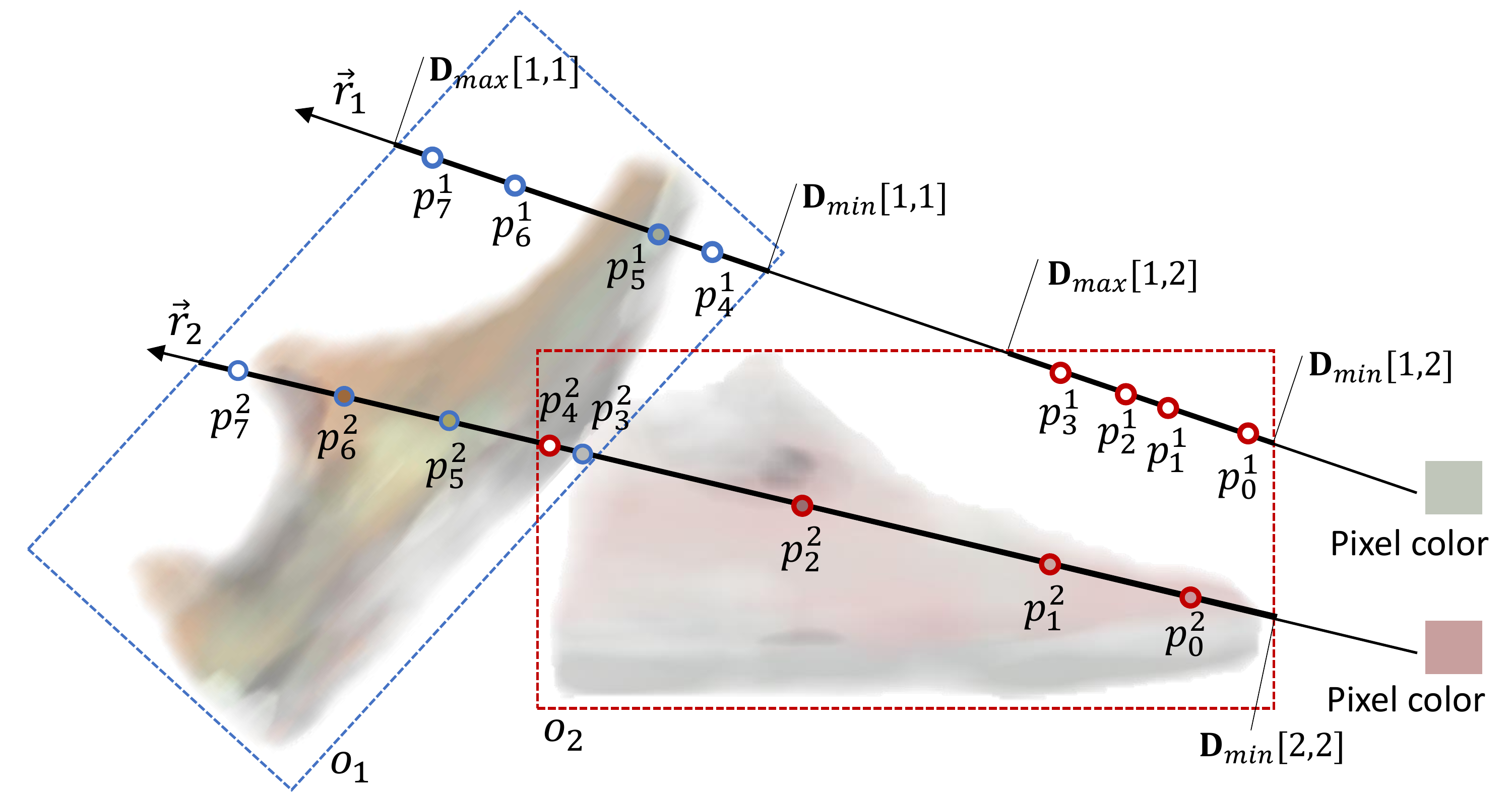}

    \caption{
        Illustration of our volumetric rendering approach with $M=2$ objects, $N=2$ rays, and $J+1 = 8$ samples. Each object is depicted by a 2D cross-section of its 3D bounding box. The colors inside each box represent the color predicted by the NeRF decoder on the cross-section (the colors in the interior of the object do not necessarily have meaning). Our raytracing step generates the matrices $\mathbf{D}_{min}$, $\mathbf{D}_{max}$ (shown), and $\mathbf{H}$ (not shown). The raymarching step generates the stratified uniform samples $(p_{j}^{n}, o_{j}^{n})$ with the object indicated by the outline color. At each sample, we query the radiance decoder for the color $c$ and density $\sigma$ at the object corresponding to the sample. We then use the volumetric rendering equation~\cite{mildenhall2020nerf} to compute the pixel color of each ray. All computations are batched.
    }
    \label{fig:rendering}
\end{figure}

Our rendering approach consists of four steps: raytracing, raymarching, radiance field query, and ray integration.
Raytracing 
uses the scene-graph, object volumes and poses to %
compute which rays may intersect with which objects.
Raymarching samples coordinates along each ray where the radiance field decoder will be queried. These radiance field queries compute density and color at each marched coordinate along the ray. 
Finally the ray integral computes the ray color as imaged by the camera.
This process is illustrated in Figure~\ref{fig:rendering}, and described in more detail in the following section.

\paragraph{Raytracing}
Given the camera pose $\camerapose$ and camera intrinsics, we generate a set of rays $R = \{\ray_{\rayidx}\}_{\rayidx=1, \dots, \numrays}$.
Each ray corresponds to a pixel in the image plane and is represented by a 3D ray origin and a direction $\dir$.
We perform a ray vs oriented-box intersection check between the $\numrays$ rays and $\numobjs$ bounding volumes $\{\volume_{\object}\}_{\object \in \scene}$.
The intersection check outputs three matrices: a hit matrix $\mathbf{H} \in {0,1}^{N \times M}$ indicating which of the $N$ rays intersect which of the $M$ objects; a minimum depth matrix $\mathbf{D}_{min} \in \mathbb{R}^{N \times M}$, where $\mathbf{D}_{min}[\rayidx,\objidx]$ is the distance where ray $\ray_{\rayidx}$ enters volume $\volume_{\objidx}$, and an analogous maximum depth matrix $\mathbf{D}_{max} \in \mathbb{R}^{N \times M}$ where $\mathbf{D}_{max}[\rayidx,\objidx]$ 
is the distance where ray $\ray_{\rayidx}$ exits $\volume_{\objidx}$.
We prune the set of rays $R$ and the rows of matrices $\mathbf{D}_{min}$, $\mathbf{D}_{max}$, $\mathbf{H}$, keeping only rays that intersect with at least one object in the scene, thereby reducing unnecessary computation.
The notation hereafter refers only to the pruned matrices.

\paragraph{Raymarching}

Our raymarcher takes as input the matrices $\mathbf{D}_{min}$, $\mathbf{D}_{max}$, $\mathbf{H}$, and outputs for each ray $\ray_{\rayidx}$ a sequence of samples $\langle (\pos_{\stepidx}^{\rayidx}, \object_{\stepidx}^{\rayidx})\rangle_{\stepidx=0 \dots \numsteps}$, where each $\pos_{\stepidx}^{\rayidx} \in \mathbb{R}^3$ is a 3D coordinate and $\object_{\stepidx}^{\rayidx}$ is an object that this sample is associated with, 
and $\numsteps+1$ refers to the number of samples along the ray.
This sequence of samples has the following properties that enable volumetric rendering:
The total number of samples is a fixed number $\numsteps+1$, which allows for subsequent batched computations, 
each object that a ray intersects with has an equal number of associated samples, 
and each sample coordinate is restricted to lie within the volume of its associated object.
The samples are sorted by depth from the ray origin.
The implementation of our ray marcher 
keeps a depth pointer $d$ and traverses along each ray starting from the smallest depth where the ray intersected an object. 
At every step, the function increments $d$ and draws a sample near $d$, 
associating it with one of the objects currently intersected.
We compute the increment step for $d$ such that when the depth pointer $d$ exits the final object, the total number of samples is $\numsteps+1$, and each object has approximately the same number of samples.

\paragraph{Radiance Field Query and Ray Integral}

The radiance decoder $\nerfmlp$ is a neural network that takes as input an object latent code $\feature$, a 3D coordinate $\pos$, and a viewing direction $\dir$, and predicts a density $\sigma$ and color $c$: $\sigma, [c] = \nerfmlp(\feature_{\object}, \pos, [\dir])$.
For each sample along the ray $\langle (\pos_{\stepidx}^{\rayidx}, \object_{\stepidx}^{\rayidx})\rangle_{\stepidx=0 \dots \numsteps}$ we look up the latent code $\feature_{\object}$ corresponding to $\object_{\stepidx}^{\rayidx}$.
We process the sequence of positions and latent codes using $\nerfmlp$ to obtain a sequence of densities and colors.
We then apply the volumetric rendering equation from NeRF~\cite{mildenhall2020nerf} to combine the densities and colors sampled along each ray $\ray$ into a single pixel color.

The inputs $\pos$ and $\dir$ to $\nerfmlp$ are expressed in the object-relative reference frame $\pose_{\object}$.
The network uses a sinusoidal positional encoding~\cite{mildenhall2020nerf} to encode $\pos$ and $\dir$.
The positional encoding of $\pos$ is concatenated with object latent code $\feature_{\object}$, before feeding into a fully-connected neural network.
The network consists of a two fully-connected backbone layers, followed by separate decoder heads for 
the color and density $\sigma$ outputs.
The density $\sigma$ does not depend on the viewing direction. %

\subsection{Predicting Grasps with the Grasp Decoder}

We use the grasp decoder $\graspmlp$ to predict stable grasps on an object $\object = (\pose_{\object}, \volume_{\object}, \feature_{\object})$ that we wish to manipulate.
The grasp prediction consists of a grasp proposal stage and a filtering stage. 
The proposal is done by a neural network $\graspmlp$ that directly consumes the latent scene representation, and the filtering relies on outputs of the radiance field described in the previous section.

\subsubsection{Grasp Decoder}
\newcommand{\graspscore}{S^{g}}
\newcommand{\grasprot}{R^{g}}
\newcommand{\grasppos}{p^{g}}
\newcommand{\gridres}{\texttt{res}}
\newcommand{\numgrasps}{K}

The grasp decoder $\graspmlp$ takes as input a 3D coordinate $\pos$ and a latent code $\feature_{\object}$ representing an object $\object$ in the scene.
It predicts a grasp score $\graspscore$ and a rotation matrix $\grasprot$.
The input coordinate $\pos$ is again in the object-relative reference frame $\pose_{\object}$.
The score $\graspscore$ models the probability that there exists a stable grasp at coordinate $\pos$.
$\grasprot$ is a $3 \times 3$ rotation matrix reflecting the grasp gripper orientation. 
Inspired by \cite{sundermeyer2021contact}, to guarantee that $\grasprot \in \mathbb{SO}(3)$ is an orthogonal matrix, or in other words, it models a pure rotation, 
the neural network outputs two 3-dimensional vectors $\mathbf{a}$ and $\hat{\mathbf{b}}$.
We assemble the rotation matrix, 
$\grasprot = [\mathbf{b} \quad \mathbf{a} \times \mathbf{b} \quad \mathbf{a}]$,
where $\mathbf{b} = (\mathbf{a} \times \hat{\mathbf{b}}) \times \mathbf{a}$,  $\mathbf{a}$ is the the gripper approach vector, and $\mathbf{b}$ is a vector that points to gripper-left.

\subsubsection{Grasp Inference and Filtering}
To predict grasps on object $\object$, we generate a set of coordinates $\langle \grasppos_i \rangle_{i}$ on a dense 3D grid of fixed resolution $res$ per axis within the volume $\volume_{\object}$.
We query $\graspmlp$ to obtain $res^{3}$ grasp proposals, each with a score $\graspscore_{i}$, rotation $\grasprot_{i}$, and the input position $\grasppos_{i}$ of which we retain the top-K proposals with the highest score.

We perform a filtering stage to reject erroneous grasp proposals:
We keep only grasps where the open gripper mesh does not collide with \emph{any} object in the scene, while the closed gripper mesh does.
We compute object-gripper collisions by querying the radiance field decoder $\nerfmlp$ across all objects.
We represent each gripper mesh as a pointcloud of 1000 randomly sampled points on the surface of the mesh.
We transform the pointclouds according to the grasp pose $(\grasppos_i, \grasprot_{i})$, and for each transformed point query the radiance field decoder for the density $\sigma$.
We keep only grasps where the total density across the 1000 points is below a threshold $T_{open}$ for the open gripper, and above a threshold $T_{closed}$ for the closed gripper.
This filtering process is possible because our object representation captures the complete object shape, rather than a view-dependent surface pointcloud.

\subsection{Learning}

\newcommand{\dataset}{\mathcal{D}}
\newcommand{\image}{I}

\subsubsection{Pre-training}
We train our model on a dataset $\dataset$ of $N_{scenes}$ scenes.
Each scene contains a ground truth scene configuration (with object volumes $\volume$ and poses $\pose$),  $N_{views}$ ground truth images rendered from random camera poses, and a set of grasp annotations consisting of grasp position $\grasppos$, orientation $\hat{\grasprot}$, and score $\hat{\graspscore}$.
Each scene has $N_{grasps}$ grasp annotations of high and low scores.
We randomly initialize a separate latent $\feature$ for each object.
The decoders $\graspmlp$ and $\nerfmlp$ are shared across all scenes and objects.
Unlike traditional NeRF, this results in learning a \emph{universal} decoder capable of decoding any object within the training distribution / category.

\begin{figure}[ht!]
    \centering
    \scriptsize
    \includegraphics[width=0.5\textwidth]{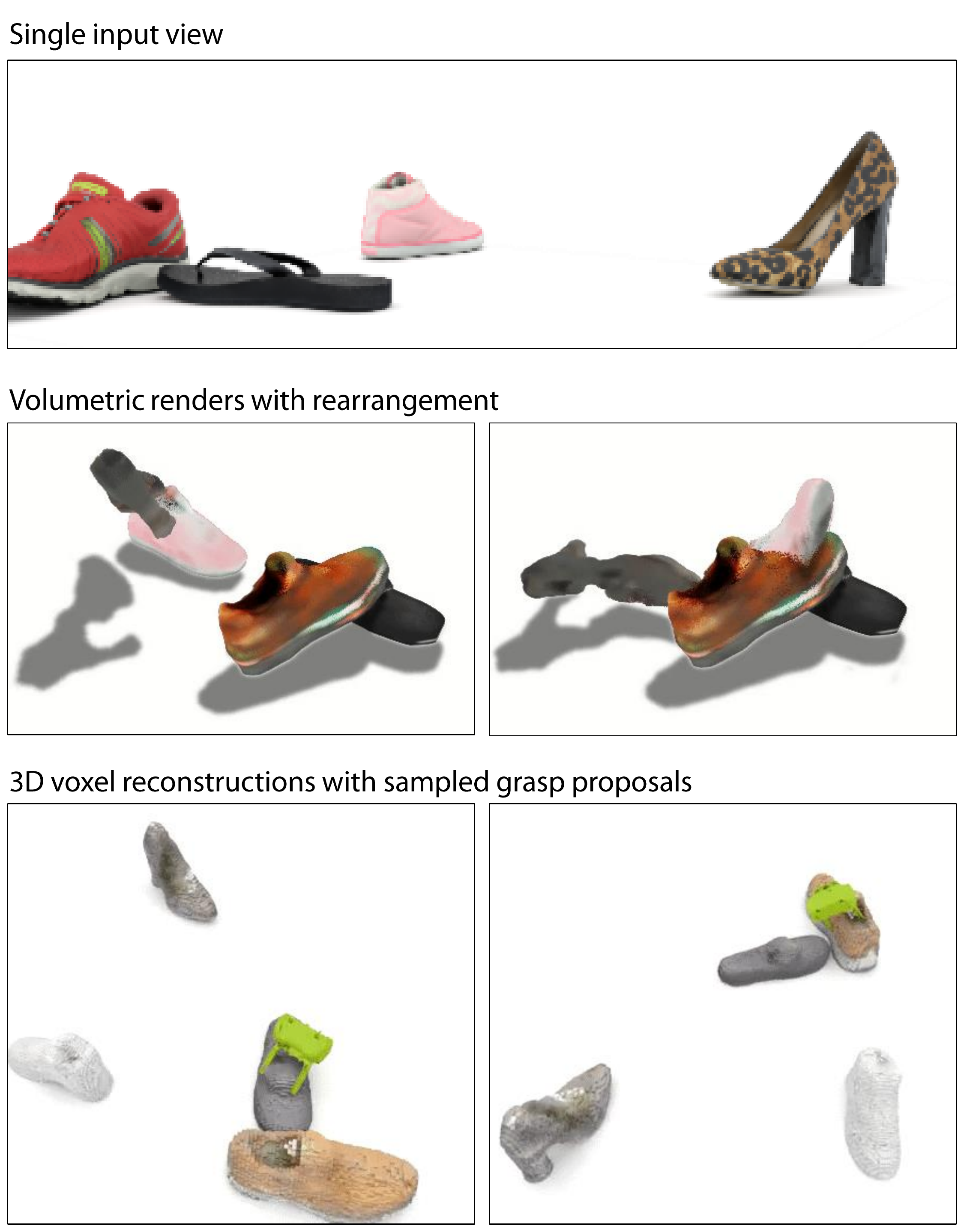}

    \caption{
        Demonstration of the scene rearrangement, compositional NeRF rendering, and 3D reconstruction capabilities of our system.
        The top row shows an RGB input view provided to the system, from which latent embeddings are obtained via inverse rendering.
        Middle row shows RGB renderings of the obtained embeddings with objects positioned in different poses from the input view.
        Notice the overlapping objects in the right image, as well as the raytraced shadows. 
        Such 3D effects are made possible by our custom raymarcher that enables raytracing the entire scene in 3D, instead of rendering each object separately and resorting to screen-space compositing. 
        Bottom row shows voxel map reconstructions extracted from our representation from two different views. 
        The voxel maps are extracted by querying the neural radiance decoder at uniform grid coordinates, and setting voxel occupancy by thresholding the predicted density. Two of the proposed grasps are illustrated in green, after performing collision-free grasp filtering with respect to all objects in the scene as well as the ground plane. 
    }
    \label{fig:rearrangement-reconstruction}
\end{figure}

We jointly optimize the NeRF reconstruction and grasp losses.
The reconstruction loss $\mathcal{L}_{\textit{rgb}}$ is the pixel-wise mean squared error between the reconstructed and ground truth images.
The grasp loss consists of two terms: grasp score loss $\mathcal{L}_{gscore}$ and rotation loss $\mathcal{L}_{grot}$.
Given the predicted grasp score and rotation $\graspscore, \grasprot = \graspmlp(\grasppos, \feature_{\object})$,
the grasp score loss is the following asymmetric least squares loss:
\begin{equation*}
    \mathcal{L}_{gscore} = \lambda \times (\textsc{ReLU}(\graspscore - \hat{\graspscore}))^2
                         + \textsc{ReLU}(\hat{\graspscore} - \graspscore)^2
\label{eq:grasp_loss}
\end{equation*}
where $\lambda \ll 1$ is a hyperparameter.
There can exist multiple grasp annotations at position $\grasppos$, some with high scores and some with low.
We want our network to predict high grasp scores if \emph{any} of the grasps at $\grasppos$ are stable, even if some grasps are not.
We achieve using the weight $\lambda$, which weighs down updates whenever the score label is lower than the predicted score.
The grasp rotation loss is the mean squared error between the predicted and ground-truth rotation, weighed by the grasp score, $\mathcal{L}_{grot} = \hat{\graspscore} (\grasprot - \hat{\grasprot}) ^ 2$.
This loss encourages the predicted rotation at position $\grasppos$ to be similar to high-scoring annotations, ignoring low-scoring ones.
During pre-training, we optimize $\mathcal{L} = \mathcal{L}_{rgb} + \mathcal{L}_{gscore} + \mathcal{L}_{grot}$ over the dataset $\mathcal{D}$ using the RMSprop optimizer.
We assume that a dataset of multiview images of objects is available,
similar to~\cite{sitzmann2019scene}, the Section~\ref{sec:dataset} describes in greater detail our proposed dataset for evaluating our method.

\subsubsection{Fine-tuning}
\label{sec:finetuning}

Given a universal decoder pre-trained as described above, when a previously unseen 
object %
is encountered, we can find a new latent code that represents this object, similar to GAN inversion~\cite{Chan2022eg3d} or inverse volumetric rendering.
We initialize a random latent code $\feature$, and optimize it using gradient descent, keeping the decoders $\graspmlp$ and $\nerfmlp$ frozen.
Note that we do not assume any grasp annotations at test time. The latent code $\feature$ is recovered by inverse volumetric rendering via optimizing the image reconstruction loss $\mathcal{L}_{rgb}$ over a \emph{single} image.

In the end, our system can be used for multiple applications, Figure~\ref{fig:rearrangement-reconstruction} shows some qualitative results.

\section{Experimental Results \label{sec:results}}

\subsection{Dataset - Shoes}\label{sec:dataset}
\begin{figure}[t]
    \centering
    \includegraphics[width=0.5\textwidth, clip, trim={0px 1300px 0px 1600px}]{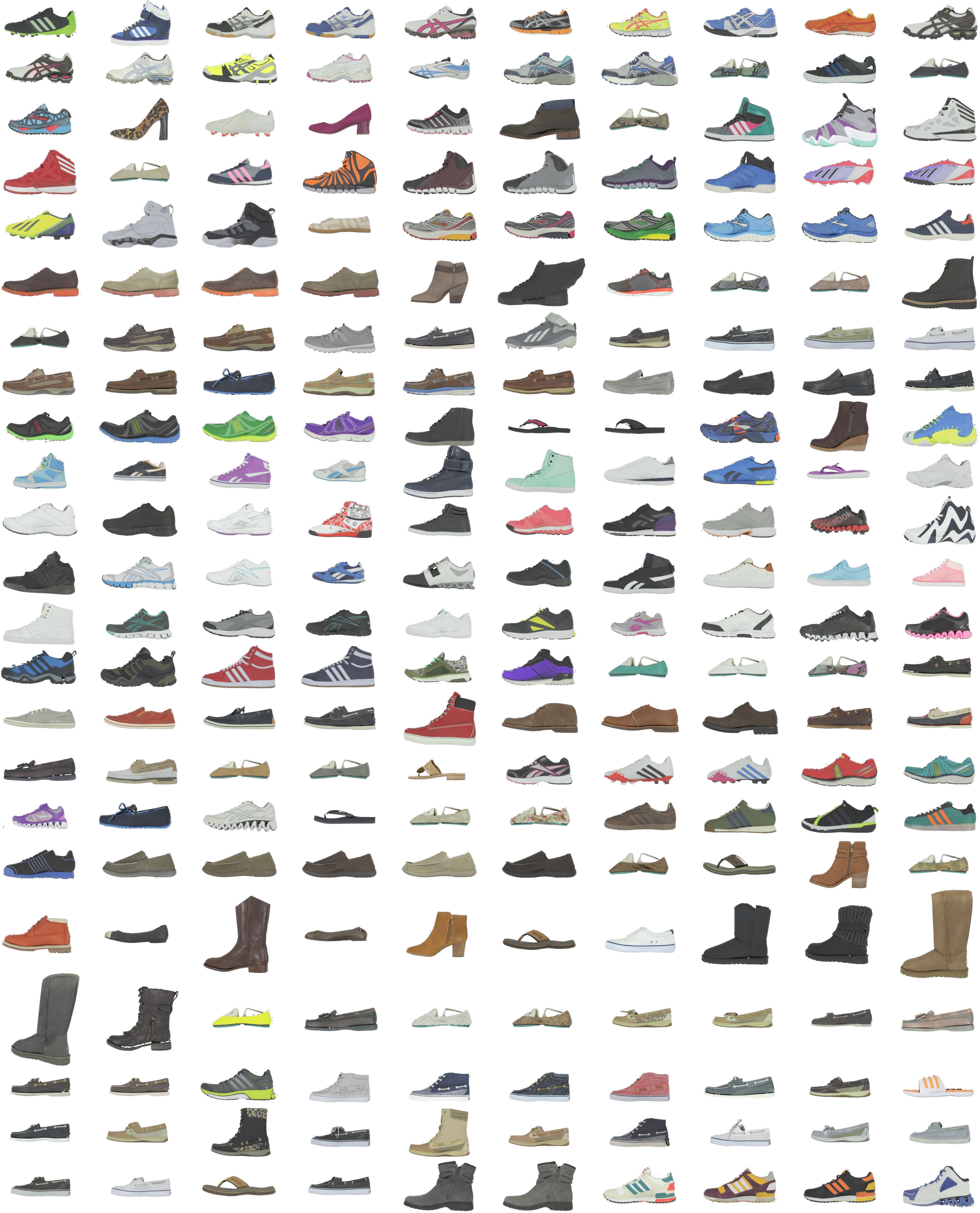}

    \caption{
    Side view render of 100 our of the 208 training shoes.
    }

    \label{fig:shoes_training}
\end{figure}

In order to evaluate our proposed method
in robotics, 
we created a suitable training dataset. This dataset comprises images of a variety of scenes %
of multiple objects within a class---in our case, shoes---arranged without self-intersections and viewed from many angles. 
The closest pre-existing dataset is composed of cars~\cite{sitzmann2019scene}.
We selected 224 different 3D models of shoes from the Google scanned dataset~\cite{downs2022google}, and ensured they all shared the same reference orientation, see Figure~\ref{fig:shoes_training} for a good summary of the training data diversity.
We split these in 208 training models, and 16 testing models.
We used the %
ray-traced renderer NViSII~\cite{morrical2021nvisii} to render 800 $\times$ 800 pixel images with 2000 samples per pixel. 
For each shoe in the dataset, we rendered 400 random views sampled from a 360-view sphere at a fixed distance (1m) from the origin, see Figure~\ref{fig:single_shoe} for examples. 
We refer to this dataset as \textbf{centered-shoe}.
Accompanying these renders, we also generated 32 scenes where three to five shoes fell onto a plane.
For each of these scenes, we randomly selected 400 views from a hemisphere at a fixed distance (1~m) from the origin of the scene, see Figure~\ref{fig:shoes_falling_test}. 
We refer to this dataset as \textbf{falling-shoes}.

\newcommand{\grasppose}{\xi^g}
\newcommand{\grasp}{g}

For each shoe within our dataset, we also include grasp poses for a 6 and 8 cm wide gripper, with quality scores. %
Inspired by \cite{wen2022catgrasp}, we sample grasp positions uniformly over the training object point cloud with approaches that are collision-free and achievable from the outside-in. 
In order to identify the grasp quality score $\graspscore \in[0,1]$ for each grasp $\grasp$, we perform $N$ trials ($N=50$ in our case) and count the number of successes in PyBullet~\cite{coumans2016pybullet}. 
Success is defined as an object staying in the gripper hand after being perturbed by random forces (under $5$~mm, $5^{\circ}$ magnitude) to its grasp pose $\grasppose = (\hat{\grasppos}, \hat{\grasprot})$ during physical simulation. %
Therefore, the grasp score can be computed as:
$$\graspscore=\frac{\sum_{i=1}^{N}\Gamma (\Delta \xi_{i} \cdot \grasppose)}{N}$$ 
where $\Gamma(\cdot)$ represents the physical simulation binary outcome about grasp success. 
$\Delta\xi_{i}$ is a random grasp pose perturbation in each trial. 
Intuitively, a robust grasp pose should be consistently stable around its neighborhood in $SE(3)$. 
This has been shown to be effective for obtaining a continuous grasp quality score~\cite{wen2022catgrasp}.

\subsection{Novel View Synthesis}

We focus our evaluation on few view synthesis. We pre-train a model on 208 shoes of the centered-shoe dataset, and we test specifically on 16 shoes. 
The scenario we want to focus on is single-view novel view synthesis. 
In greater detail, a pre-trained model is allowed to see either a single view of the object, and then it is asked 
to re-render that object in different camera poses. 
See Figure~\ref{fig:single_shoe} for a single view example.
With respect to the falling-shoe dataset, we explore the same scenarios, made more challenging due to the partially occluded views of some shoes. %
We also propose a few-view scene editing task. Given a model trained on the centered-shoe dataset, the task is to render synthetic scenes with novel arrangements of specific shoes. The shoes chosen to be rendered are identified from single RGB image. First, we optimize a latent code for each chosen shoe from the provided single view image. Then, given a scene configuration of poses, we render views of this new scene from any viewpoint. We quantitatively test our model, pre-trained on the centered-shoe dataset, using the scenes from the falling-shoe dataset.

For evaluation 
we use the standard NeRF image quality metrics~\cite{mildenhall2020nerf}; Peak Signal to Noise Ratio (PSNR), Structural
Similarity Index Measure (SSIM)~\cite{wang2004image}, and Learned Perceptual Image Patch Similarity (LPIPS)~\cite{zhang2018unreasonable}.
We evaluate the methods on 75 randomly selected views for each scene.

\subsection{Results on synthetic data of shoes}

In our final proposed method, we leverage latent code and decoder optimization to find the best fitting NeRF model for our scene. 
In these experiments, we are interested in understanding this design choice; as such, we design 3 different optimizations, 
1) only the latent code is optimized, 
2) only the decoder is optimized, whereas a random latent code is sampled, 
3) both latent and decoder are optimized.

\begin{table*}[!htp]\centering

\scriptsize
\begin{tabular}{l|cccc|cccc|cccc}\toprule
&\multicolumn{4}{c|}{PSNR ↑} &\multicolumn{4}{c|}{SSIM ↑} &\multicolumn{4}{c}{LPIPS ↓} \\\midrule
&pnerf\cite{yu2021pixelnerf}&latent &model &both &pnerf\cite{yu2021pixelnerf}&latent &model &both &pnerf\cite{yu2021pixelnerf} &latent &model &both \\
color           & 24.44 & 24.05 & 23.64 & 23.66 & 0.903 & 0.908 & 0.902 & 0.903 & 0.125 & 0.103 & 0.100 & 0.100 \\
shape           & 21.35 & 21.10 & 20.50 & 20.43 & 0.883 & 0.885 & 0.875 & 0.875 & 0.140 & 0.123 & 0.121 & 0.122 \\
shape and color & 25.22 & 25.28 & 24.58 & 24.63 & 0.912 & 0.916 & 0.916 & 0.916 & 0.117 & 0.093 & 0.081 & 0.081 \\
challenge       & 22.91 & 21.77 & 21.61 & 21.59 & 0.904 & 0.900 & 0.894 & 0.895 & 0.124 & 0.113 & 0.109 & 0.109 \\
\midrule
& \textbf{23.48} & 23.05 & 22.58 & 22.58 & 0.900 & \textbf{0.902} & 0.897 & 0.897 & 0.126 & 0.108 & \textbf{0.103} & \textbf{0.103} \\
\bottomrule
\end{tabular}
\vspace{-1em}
\caption{Center-Shoe Dataset Results}
\vspace{-1em}

\label{tab:results_synthetic_data}
\end{table*}

\textbf{Centered-shoe} - 
Table~\ref{tab:results_synthetic_data} shows different RGB-based reconstruction metrics after optimizing for 100 epochs (taking about 1 minute).
We report the best optimization performance taken from 15 randomly selected viewpoints (which are shared across all experiments). 
We compare to pixelNeRF (pnerf)~\cite{yu2021pixelnerf} as a baseline, using the same paradigm and images for training. We use
their released source code. 
Comparing the methods, pnerf has over 28 million parameters,%
and needs 1.25 seconds to render a $128 \times 128$ image.
On the other hand, when doing novel view synthesis, our method only needs 40k parameters and can render the same $128 \times 128$ frame in 20 millisecond. This does not include the grasping head, which adds 30k parameters.

\begin{figure}[h]
    \centering
    \scriptsize
    \begin{tabular}{c|cccc}
        gt & pnerf\cite{yu2021pixelnerf} & latent & decoder & both \\
        \hline 
        
        \includegraphics[clip,trim={70px 70px 70px 70px},width=0.14\linewidth]{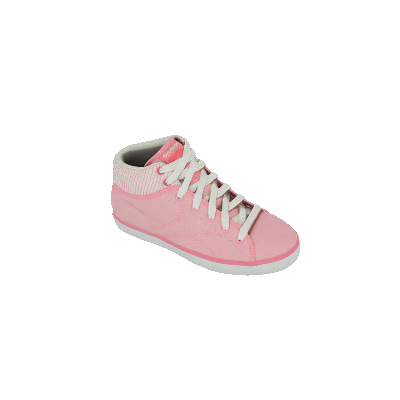} &
        \includegraphics[clip,trim={22px 22px 22px 22px},width=0.14\linewidth]{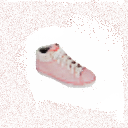} &
        
        \includegraphics[clip,trim={70px 70px 70px 70px},width=0.14\linewidth]{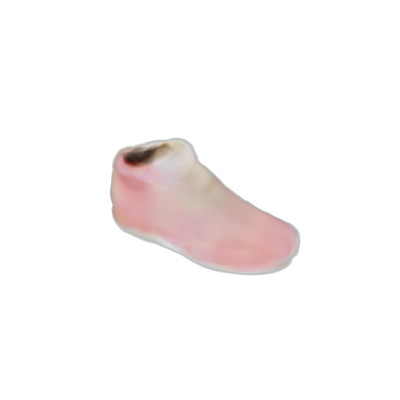} &
        \includegraphics[clip,trim={70px 70px 70px 70px},width=0.14\linewidth]{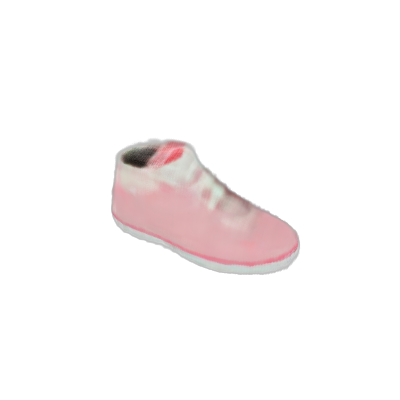} &
        \includegraphics[clip,trim={70px 70px 70px 70px},width=0.14\linewidth]{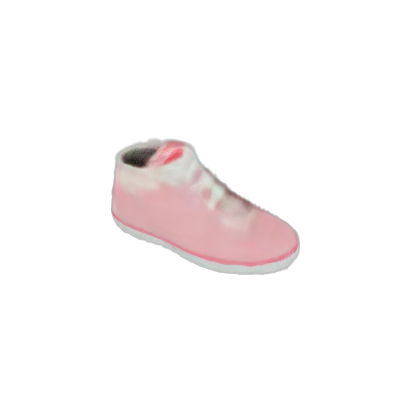} \\

        \includegraphics[clip,trim={70px 70px 70px 70px},width=0.14\linewidth]{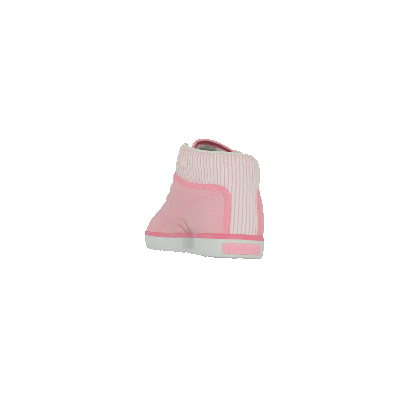} &
        \includegraphics[clip,trim={22px 22px 22px 22px},width=0.14\linewidth]{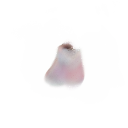} &
        
        \includegraphics[clip,trim={70px 70px 70px 70px},width=0.14\linewidth]{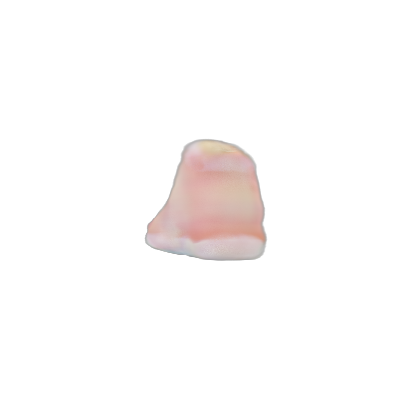} &
        \includegraphics[clip,trim={70px 70px 70px 70px},width=0.14\linewidth]{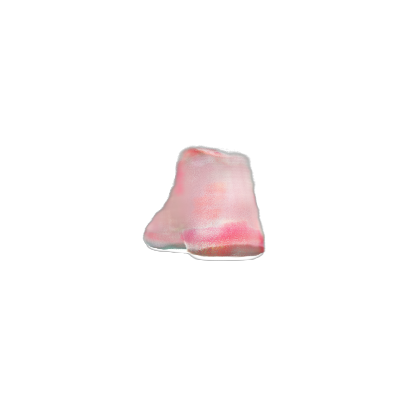} &
        \includegraphics[clip,trim={70px 70px 70px 70px},width=0.14\linewidth]{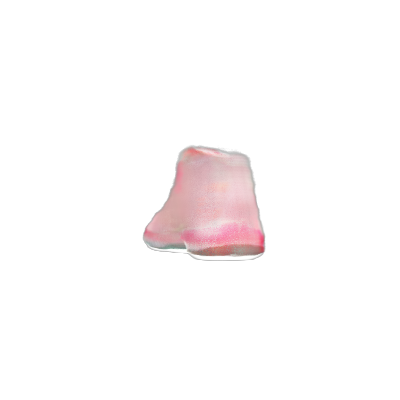} \\

        \includegraphics[clip,trim={22px 22px 22px 22px},width=0.14\linewidth]{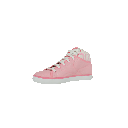} &
        \includegraphics[clip,trim={22px 22px 22px 22px},width=0.14\linewidth]{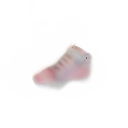} &
        \includegraphics[clip,trim={70px 70px 70px 70px},width=0.14\linewidth]{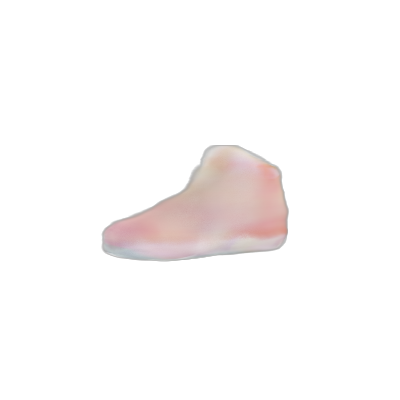} &
        \includegraphics[clip,trim={22px 22px 22px 22px},width=0.14\linewidth]{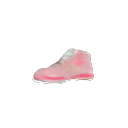} &
        \includegraphics[clip,trim={22px 22px 22px 22px},width=0.14\linewidth]{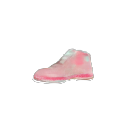}\\       

        \includegraphics[clip,trim={22px 22px 22px 22px},width=0.14\linewidth]{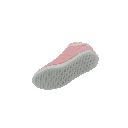} &
        \includegraphics[clip,trim={22px 22px 22px 22px},width=0.14\linewidth]{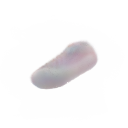} &
        \includegraphics[clip,trim={70px 70px 70px 70px},width=0.14\linewidth]{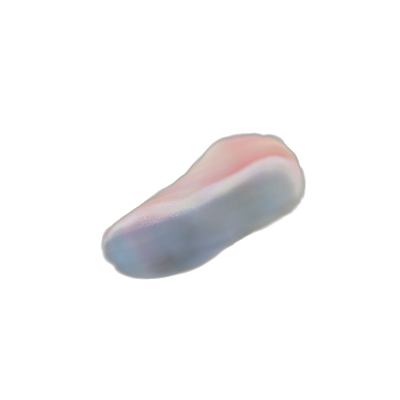} &
        \includegraphics[clip,trim={22px 22px 22px 22px},width=0.14\linewidth]{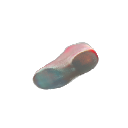} &
        \includegraphics[clip,trim={22px 22px 22px 22px},width=0.14\linewidth]{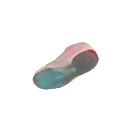}\\  

    \end{tabular}
    \vspace{-1em}
    \caption{The first row shows the input view, while the other rows show synthesized novel views not available to the methods. Our method retains the semantics of the object in the different views. 
    }
    \label{fig:single_shoe}
\end{figure}

Even though our proposed method is multiple orders of magnitude smaller than pnerf, we achieve similar results.
Our method outperforms pnerf on LPIPS, which measures semantic similarity to ground truth, while we are almost the same on SSIM and slightly worse on PSNR. Pnerf, similar to our method, has to also be pre-trained on category-level data.
Figure~\ref{fig:single_shoe} shows qualitative results on the selected single input view for a test object (Reebok SH PRIME COURT MID) and its reconstruction counterpart, as well as a synthesized novel view. 
The qualitative results suggest that (1) our smaller architecture might be achieving better generalization, while sacrificing high-frequency details (e.g. the shoelaces), and (2) pnerf gives more detailed results from novel views close to the input view, but more blurry results from views that are very different.
Optimizing both the latent code and decoder allows for better reconstruction performance according to LPIPS, but worse on PSNR and SSIM compared to only optimizing the latent code.

\textbf{Falling-Shoe and Scene Editing} - 
We also evaluate our system in a complex environment where three to five 
test shoes are randomly placed on a tabletop. 
For this experiment we only looked at single image optimization. 
Similar to the previous experiment, optimizing both the decoder and latent code shows greatest performance, with a
PSNR of 19.84, 20.41, and 20.78 for the latent, decoder, and both, respectively.
We observe a PSNR of 20.94, which is better than doing single view optimization, although on the other metrics, it performed worse, 
{\em e.g.}, 0.114 {\em vs.} 0.105 (lower is better) for LPIPS metric.

\begin{figure*}
    \centering
     \includegraphics[width=\textwidth]{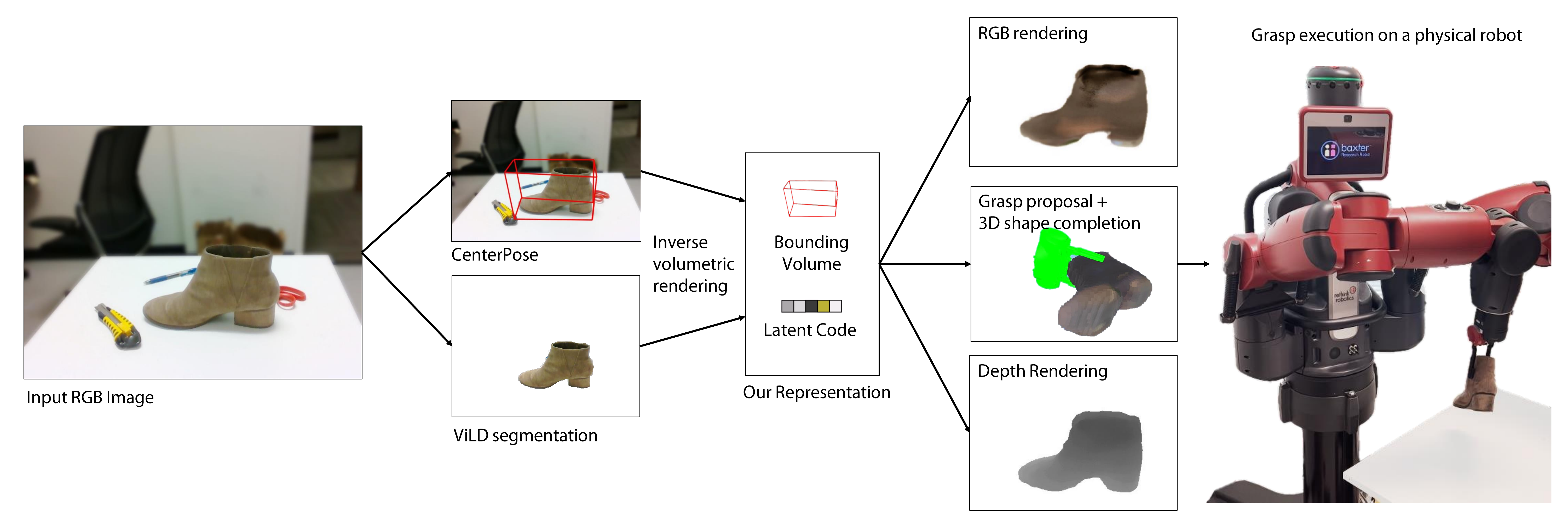}
     \caption{Our system consumes an RGB image calibrated with respect to the robot. 
     We retrieve the object pose and dimensions from the image using CenterPose~\cite{lin2022icracenterpose}. 
     We leverage a segmentation algorithm (ViLD)~\cite{gu2022openvocabulary} to mask out rays that do not intersect the object. 
     We then perform inverse volumetric rendering to obtain a latent representation of the object.
     We can then use this representation to render RGB or depth images of the object, generate stable grasp proposals, and obtain a 3D reconstruction with shape completion.
     The proposed grasps can be executed on a physical robot.
     }
     \label{fig:inference}
\end{figure*}

\begin{figure}
    \centering
      \includegraphics[width=0.5\textwidth]{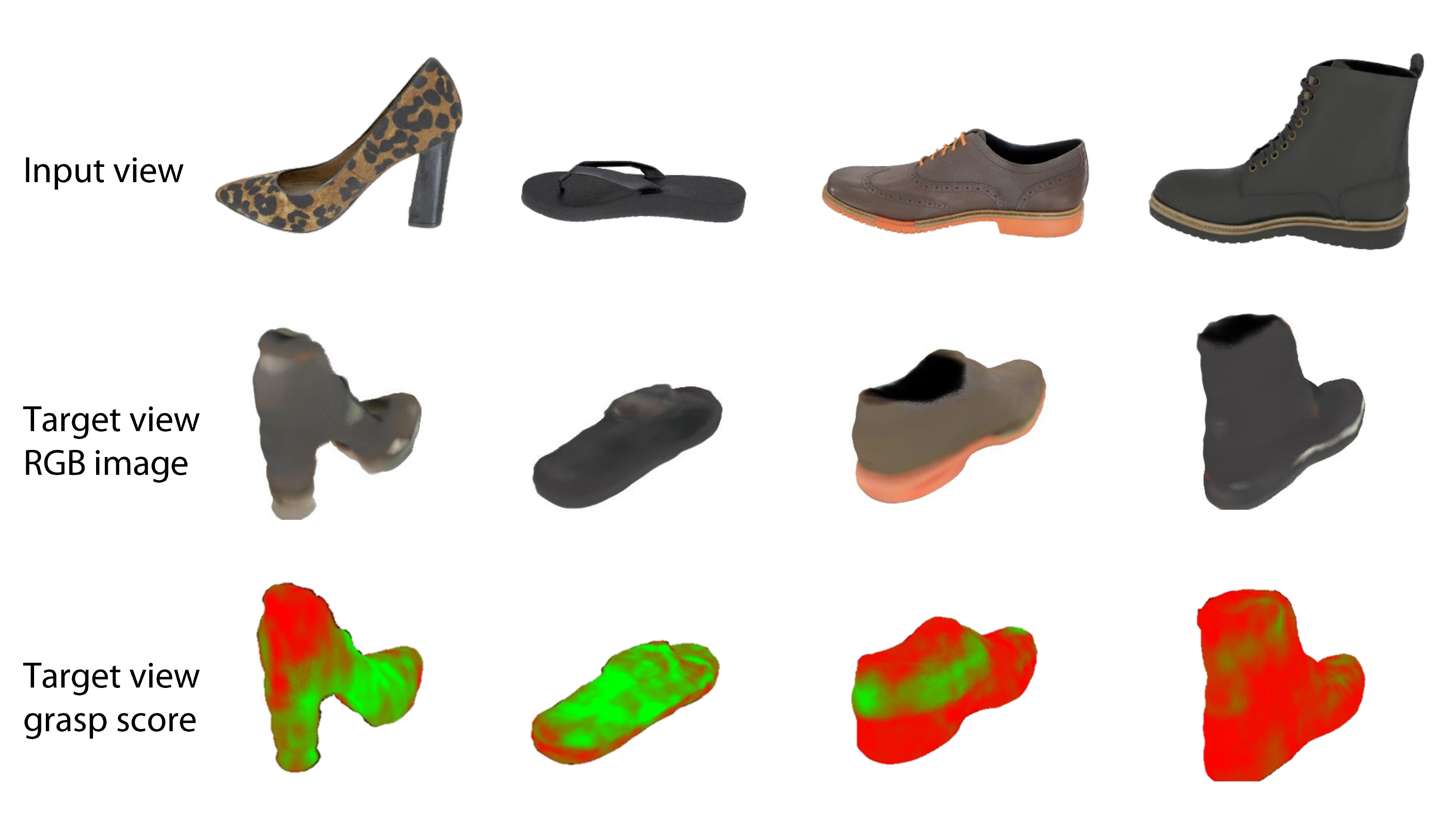}
     \caption{
        Visualizations of renderings and predicted grasp score fields on diverse unseen objects, illustrating how our system responds to different object shapes and sizes.
        First row shows the input view from which we compute a latent representation.
        Second row shows a novel view rendering from our radiance field decoder.
        Third row shows a novel view rendering, where the radiance color is replaced with a color-mapped grasp score for a Baxter gripper. Red corresponds to grasp score 0 and green corresponds to grasp score 1.
        The boot in the rightmost column is too big to fit in the gripper, thus only a small area is deemed graspable. The slipper in the second column is narrow enough to fit in the gripper. Our system has not been exposed to grasp annotations for these objects.
     }
     \label{fig:graspscore}
     \vspace{-2em}
\end{figure}

\subsection{Grasping and Robotics Experiments}

\begingroup

Following the same split for the synthetic shoe dataset, we report the grasping success rate.
For each testing image we perform inverse volumetric rendering through the NeRF decoder $\nerfmlp$ to recover a latent representation, and use it to predict the 5 most confident grasps using the grasp decoder $\graspmlp$.
Note that only the radiance field decoder $\nerfmlp$ participates in the latent code optimization by providing gradients to the latent code, as there are no grasp annotations available at test-time. 
Both the grasp decoder $\graspmlp$ and NeRF decoder $\nerfmlp$ remain frozen.
As baseline we use GraspNet~\cite{mousavian2019graspnet,pytorch_6dof-graspnet} to generate grasps from 
the partial view point cloud. 
The input to our system is an RGB image. The input to GraspNet is a partial-view pointcloud captured from the same perspective as the RGB image. Thus this is not an even comparison, as GraspNet has access to depth information.
We evaluate grasp success using the PyBullet physics simulator \cite{coumans2016pybullet}.

We report a success rate for top-1 grasp prediction of 70.2\% for GraspNet {\em vs.} 61.6\% for our proposed method. 
While using top-5 decreases the performance of GraspNet to 69.7\%, our method improves to 
64.2\%. 
Top-5 refers to the percentage that are correct within the top-5 predictions, {\em e.g.}, 
if 3 out of 5 grasps are correct, the method gets awarded a score of 60\%, not to confound with top-5 prediction in computer vision (where a full score is awarded if the right prediction is in the top-5). 
Even though our performance is slightly lower than GraspNet, our method still achieves useful results, even though it was trained on a relatively small category-level dataset, and had access to only RGB input.
Figure~\ref{fig:graspscore} shows qualitative results showing our predicted grasp score fields on a variety of objects.
Our goal in this paper is not to achieve state-of-the-art grasping performance - for that we would still recommend the reader to use systems specially designed for grasping and leverage depth cameras.
Our main finding that we wish to communicate is that a latent representation recovered by inverting a category-level pre-trained neural radiance field is useful for solving a robotics task unrelated to novel-view synthesis.
In addition, the resulting representation facilitates multiple downstream uses, including 3D reconstruction with shape completion, novel view synthesis, and the collision checking that we used to propose collision-free grasps.
This finding opens up interesting future research towards streamlining robotics representations, and using such latent representations as backbones for robot manipulator systems.

\vspace{-5mm}
\section{Conclusion}
We presented a unified object-centric implicit representation that can be used
for RGB and depth novel view rendering, 3D reconstruction, and proposing stable grasps.
Our main insight is that a latent object representation obtained by inverse rendering of a pre-trained neural radiance field can be leveraged for robotics tasks with reasonably good performance (see Figure~\ref{fig:inference}).
Our method does not use an encoder network, which reduces the need for diverse, domain-randomized training data.
Instead, we train only a decoder network on a clean and lightweight synthetic multiview dataset, which we invert at test-time to handle variations like partial views or to generalize to real images.

\bibliographystyle{IEEEtran}
\bibliography{main.bib}

\end{document}